\documentclass[runningheads]{llncs}
\usepackage{graphicx}
\usepackage{subfigure}
\usepackage{pifont}
\usepackage{color}
\usepackage{longtable}
\usepackage{multirow}
\usepackage{enumitem}

\begin{document}
\title{Data-Algorithm-Architecture Co-Optimization for Fair Neural Networks on Skin Lesion Dataset}
\titlerunning{Data-Algorithm-Architecture Co-Optimization for Fair Neural Networks}

\author{Yi Sheng\inst{1} \and
Junhuan Yang\inst{1} \and
Jinyang Li\inst{1} \and
James Alaina\inst{2} \and
Xiaowei Xu\inst{3} \and
Yiyu Shi\inst{4} \and
Jingtong Hu\inst{5} \and
Weiwen Jiang\inst{1} \and
Lei Yang\inst{1}
}
\authorrunning{Y. Sheng et al.}
%
\institute{George Mason University \\
\email{ysheng2@gmu.edu, lyang29@gmu.edu}\\
\and
 University of Pittsburgh Medical Center \and
 Guangdong Provincial People's Hospital \and
 University of Notre Dame \and
 University of Pittsburgh
}

%
%
%
\maketitle              
\begin{abstract}

As Artificial Intelligence (AI) increasingly integrates into our daily lives, fairness has emerged as a critical concern, particularly in medical AI, where datasets often reflect inherent biases due to social factors like the underrepresentation of marginalized communities and socioeconomic barriers to data collection. Traditional approaches to mitigating these biases have focused on data augmentation and the development of fairness-aware training algorithms. However, this paper argues that the architecture of neural networks, a core component of Machine Learning (ML), plays a crucial role in ensuring fairness. We demonstrate that addressing fairness effectively requires a holistic approach that simultaneously considers data, algorithms, and architecture. Utilizing Automated ML (AutoML) technology, specifically Neural Architecture Search (NAS), we introduce a novel framework, BiaslessNAS, designed to achieve fair outcomes in analyzing skin lesion datasets. BiaslessNAS incorporates fairness considerations at every stage of the NAS process, leading to the identification of neural networks that are not only more accurate but also significantly fairer. Our experiments show that BiaslessNAS achieves a 2.55\% increase in accuracy and a 65.50\% improvement in fairness compared to traditional NAS methods, underscoring the importance of integrating fairness into neural network architecture for better outcomes in medical AI applications.

\end{abstract}
\section{Introduction}

\begin{figure}[t]
  \centering

  \includegraphics[width=1\textwidth]{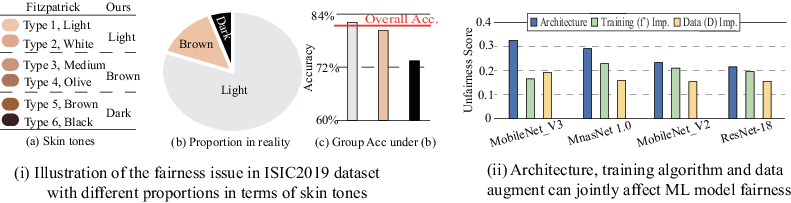}
    
    \caption{Bias issue behind training dataset and three fairness-related factors
    }
    
  \label{fig:task-mot}

\end{figure}

The democratization of AI is rapidly expanding the use of machine learning, notably neural networks, across various medical disciplines \cite{wang2020ica,zheng2021hierarchical}, with dermatology leading due to the availability of comprehensive skin lesion datasets \cite{de2020use}. However, unlike general-purpose image datasets like ImageNet \cite{krizhevsky2012imagenet}, skin lesion datasets often exhibit biases, particularly regarding skin tone. This imbalance poses a significant challenge for machine learning in dermatology, as it can result in models that, while accurate on average, perform poorly for underrepresented groups. Our analysis of the ISIC2019 dermatology dataset \cite{ISIC2019} revealed a notable accuracy disparity of over 10\% between lighter and darker skin tones, despite an overall accuracy of 81.71\% in Fig. \ref{fig:task-mot}(i). This issue of skin-type bias is not unique to academic datasets but is also prevalent in commercial AI applications, including facial-analysis tools \cite{biasapp} and Skin Image Search platforms \cite{kamulegeya2019using}.


Researches \cite{nakajima2019generating,miranda2022debiasing} have highlighted that data bias significantly impacts the fairness of machine learning (ML) models. And
Fig. \ref{fig:task-mot}(ii) shows that except data, algorithm and network also
affect the fairness, and one observation from Table \ref{tab:table3} shows that co-optimization of these factors yields the best performance.
Through a comprehensive review of the ML process, we've found that neural architectures and training algorithms, alongside data, also influence fairness. Interestingly, these factors are interconnected, suggesting that optimizing them in isolation may not yield the most equitable outcomes. While previous studies have focused on enhancing fairness from data \cite{spinde2021towards,9488320} or algorithmic \cite{bahng2020learning,nam2020learning,chiu2023toward,jin2024learning,peng2024maxk} perspectives, the role of neural architecture remains underexplored. Neural Architecture Search (NAS), which has gained attention for improving model performance and efficiency \cite{jiang2019accuracy,jiang2020hardware,peng2024lingcn,peng2023autorep}, involves search space formulation, architecture evaluation, and optimizer evolution. This process offers a unique avenue to integrate data processing, training algorithms, and architecture search within a unified framework, yet fairness considerations have largely been overlooked in NAS, especially regarding biomedical data.

In response, this paper introduces Biasless-NAS, a comprehensive framework that leverages NAS for the co-optimization of data, training algorithms, and neural architecture. BiaslessNAS embeds fairness awareness into each phase of the NAS process, ensuring that these elements are simultaneously optimized for fairness in skin lesion dataset analysis. This approach not only addresses the gap in incorporating fairness into NAS but also sets a new standard for developing equitable ML models in biomedical applications.
Experimental results show that BiaslessNAS can achieve the highest accuracy with a fairness improvement of 33.13\%. 
With tolerant accuracy degradation, BiaslessNAS can find a fairer neural architecture with 65.59\% fairness
improvements.

\section{Related Work}
With the biased data in hand,
traditional approaches can be divided into two directions: (1) data bias removal, and (2) fair training.
Data bias removal: one way to remove the bias is by building a balanced dataset, however, it is a time-consuming process.
An alternative way is to employ data augmentation.
For example, \cite{abusitta2019generative} generates biased sets to increase the minority data artificially. 
In addition to data balance \cite{hao2021towards,sharma2020data}, techniques were proposed to modify the training algorithms in addressing the fairness issue.
Authors in \cite{shafahi2019adversarial,li2021estimating} applied adversarial training and add a discrimination module to improve fairness. 

Our work stands at a different point to consider the neural architecture in addressing the fairness issue. We propose a framework to jointly optimize neural architectures, training algorithms, and data augmentation. 
The above-mentioned debiasing methods can be integrated into our framework.

\section{Method}

\subsection{Fairness Metric Definition and Factor Investigation}
\noindent
Given a neural architecture $N$ and datasets $\langle T,D\rangle$ where $T$ is the training set and $D$ is the validation set, $N$ is trained on $T$ to generate the model $f^\prime_N$, which is then validated on $D$ to obtain accuracy $A(f_N^{\prime},D)$.
Fairness exists because data in $D$ have additional attributes (e.g., skin tones), which will divide $D$ into groups, denoted $\{D_{g_1}, D_{g_2}, \cdots, D_{g_K}\}$.
For example, if a dataset contains two skin tones (i.e.,  $g_1=light\_skin$ and $g_2=dark\_skin$), the accuracy of model $f_N^{\prime}$ on group $g_i$ is denoted as $A(f_N^{\prime},D_{g_i})$.

We define the ``unfairness score'' based on the overall accuracy and the group accuracy, denoted as $U(f_N^\prime,D)$.
Specifically, in this project, we calculate the unfairness score \cite{li2021estimating} $U(f_N^\prime,D)$ as the L1-norm:
\begin{equation}\label{equ:fair}
\small
    U(f_N^\prime,D)=\sum_{\forall g_i\in G}\{|A(f_N^{\prime},D_{g_i})-A(f_N^\prime,D)|\}.    
\end{equation}

Results in Fig \ref{fig:task-mot} (ii) illustrate that different architectures ($N$) have different unfairness scores.
We further investigate the influence of the training approach and data preprocessing.
In Fig. \ref{fig:task-mot} (ii), we modify the loss function in training to consider fairness in the training process, denoted as ``Training Imp.'', and we conduct data balancing to increase the samples in minority groups aiming at improving fairness, denoted as ``Data Imp.''.
It is clear that both approaches can reduce the unfairness score.
More interestingly, the three factors $N$, $f^{\prime}$, and $D$ are coupled with each other, which 
indicates that optimizing them simultaneously is best to minimize the unfairness score.

\subsection{BiaslessNAS Framework}

\begin{figure*}[t]
  \centering
  
  \includegraphics[width=1\textwidth]{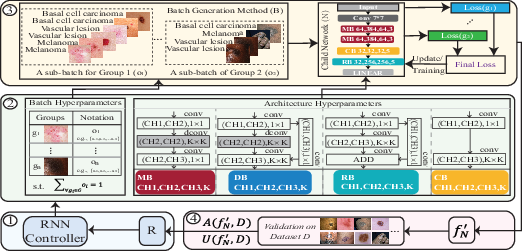}
  \caption{Overview of BiaslessNAS
  : \ding{192} controller: generating a reward $R$ and updating the recurrent neural network (RNN)-based controller; \ding{193} search space: sampling a set of hyperparameters based on the updated controller to obtain the batch composition of groups' data and a child network; \ding{194} fairness-aware trainer: on a validated dataset, training the identified child network on the generated batches; \ding{195} evaluator: generate the accuracy and unfairness score for the trained neural network $f_N^{\prime}$.
  }
    
  \label{fig:overviewframework}
  
\end{figure*}

\noindent
\textbf{Overview of BiaslessNAS framework:} Fig. \ref{fig:overviewframework} shows the overview of BiaslessNAS, which is composed of 4 components: \ding{192} \textit{reinforcement learning (RL) optimizer}, {\ding{193}} \textit{search space}, {\ding{194}} \textit{fairness-aware trainer}, and {\ding{195}} \textit{fairness and accuracy evaluator}.
Specifically, a recurrent neural network (RNN)-based controller guides the optimization process by sampling a batch generation method ($BGM$) and a neural architecture (a.k.a., child network) $N$ in the \textit{search space}. 
Then, \textit{the fairness-aware trainer} will tune the child network.
Next, in the \textit{evaluator}, the obtained model from the trainer will be evaluated to obtain accuracy and unfairness scores.
With these metrics, a reward will be generated, which will be used to update RNN in the controller.
We will introduce the details of these components in the following texts.

\noindent{\large\ding{192}} \textbf{RL Optimizer}: The controller iteratively predicts the hyperparameters of the batch generation method $BGM$ and the child network $N$. 
In each iteration, the controller receives a reward to update the RNN network. The reward $R$ is generated based on the outputs of the evaluator (see \ding{195}), including accuracy $A(f_N^{\prime},D)$, and unfairness score $U(f_N^\prime,D)$.
$R$ is computed as follows.

\begin{equation}\label{equ:reward}
\small
R = \left\{ {\begin{array}{*{20}{c}}
{\alpha\cdot A(f_N^{\prime},D)-\beta\cdot U(f_N^\prime,D)}&{\ \ \ A(f_N^{\prime},D)\ge AC}\\
{-1}&{otherwise}
\end{array}} \right.
\end{equation}
where $\alpha$, $\beta$ are two scaling factors that could be adjusted according to the specific
demands on accuracy or fairness, and $AC$ is the requirement of the model accuracy on the full dataset $D$.

Based on the reward, we employ reinforcement learning to update the controller.
Specifically, we apply the Monte Carlo policy gradient algorithm \cite{williams1992simple}:

\begin{equation}
\small
    \nabla J(\theta) = \frac{1}{m}\sum_{k=1}^{m}\sum_{t=1}^{T}\gamma^{T-t}\nabla_{\theta}\log \pi_{\theta} (a_{t}|a_{(t-1):1})(R_{k}-b)
\end{equation}
where $m$ and $T$ are the batch size and step in each episode. Rewards are discounted by
an exponential factor $\gamma$, and $b$ is the average exponential moving.

\noindent{\large\ding{193}} \textbf{Data/Architecture Fusing Search Space}: The search space is composed of two sets of hyperparameters: (1) hyperparameters for $BGM$, and (2) hyperparameters for child network architecture $N$.

\textit{Batch Generation.} The idea of creating $BGM$ is to adjust the composition of data from different groups in one training data batch.
We define $o_i$ to be a ratio, indicating the percentage of images in one batch comes from sub-dataset $D_{g_i}$.
Let $BS$ be the batch size, then, we have $o_i\times BS$ to be the number of images from sub-dataset $D_{g_i}$, and we have the constraint that $\sum_{\forall g_i \in G}\{o_i\}=1$.
To avoid accuracy degradation caused by oversampling of minority groups, we additionally have the following constraint: $\forall g_i \in G, g_j \in G$, if $|D_{g_i}|\le |D_{g_j}|$, then $o_i\le o_j$, where $|D_{g_k}|$ indicates the size of sub-dataset $D_{g_k}$.

\textit{Neural Architecture.} We apply a linear array of a block as the backbone architecture.
The design of basic blocks is inspired by the existing popular convolutional neural networks, including VGG-Net \cite{simonyan2014very}, MobileNet \cite{howard2017mobilenets}, and ResNet \cite{he2016deep}.
In this work, as shown in Fig. \ref{fig:overviewframework} \ding{193}, we involve four types of basic blocks, including MobileNetV2-inspired ones (i.e., MB and DB), ResNet-inspired block (RB), and VGG-inspired block (CB).
The basic blocks have four hyperparameters, including channel numbers ($CH1$, $CH2$, and $CH3$) and kernel sizes ($K$).
Kindly note that $CH1$ is not a searchable hyperparameter.
Considering two adjacent blocks ($A_i\rightarrow A_j$), $CH1$ in $A_j$ has the same value as $CH3$ in $A_i$.
Besides the four types of blocks, we also enable the block to be a skip operation, so that it has the flexibility in searching for the depth of the neural network.

\noindent{\large\ding{194}} \textbf{Fairness-aware Trainer}: Given an identified architecture (i.e., child network $N$) and $BGM$, the fairness-aware trainer trains the child network to generate a trained model $f_N^{\prime}$.
Specifically, we first create batches of data using $BGM$ on the validation dataset.
Then, the model is trained using a fairness-aware loss function.
Finally, after the iterative training process, we can obtain $f_N^{\prime}$. 

Particularly, the fairness-aware loss function is formulated by leveraging the hyperparameters in $BGM$.
Denote $B_{g_i}$ as the sub-batch of samples from sub-dataset $D_{g_i}$, and we have $|B_{g_i}|=o_i\times |D_{g_i}|$, where $|*|$ is the size of a dataset/batch, and $o_i$ is the ratio in $BGM$.
For each sample $s\in B_{g_i}$, it has a target label $T_s$ and a prediction results $P_s$. 
After the forward propagation, we apply Cross Entropy to compute the loss, as follows,

\begin{equation}
\small
    L = -\sum_{g_i\in G}\sum_{s\in B_{g_i}}\{\frac{\arg\max_{g_j\in G}{o_j}}{o_i} \cdot T_s \log P_s\},
\end{equation}

where $\arg\max_{g_j\in G}{o_j}$ identifies the ratio of the largest group to compose a batch, which is used to form a scaling factor. 
The final generated fair loss is also used to complete the backward propagation.

\noindent{\large\ding{195}} \textbf{Fairness and Accuracy Evaluator}:  With the trained model $f_{N}^{\prime}$, the accuracy can be obtained. Meanwhile, the unfairness score can be calculated based on the validate dataset $D$ with Eq. \ref{equ:fair}.
The obtained $A(f_N^{\prime},D)$ and $U(f_N^\prime,D)$ will be utilized to calculate the reward in \ding{192} RL Optimizer.

\section{Experiment}


\noindent
\textbf{\textit{Dataset and settings}}
We use the Fair and Intelligent Embedded System Challenge (ESFair) dataset \cite{esfair}, which is composed of data from ISIC2019, Dermnet\cite{Dermnet}, and Atlas\cite{Altas}. Thera are 5 dermatology diseases for classification.
We compare solutions obtained by BiaslessNAS with a set of existing neural architectures, including MobileNetV2 \cite{sandler2018mobilenetv2}, ResNet \cite{targ2016resnet}, and MnasNet \cite{tan2019mnasnet}.
All models are trained from scratch with the same hyperparameters on a GPU cluster with 48 RTX 3080.
The learning rate starts from 0.01 with a decay of 0.9 in 20 steps; while the batch size is 32 with 500 epochs.



\begin{table*}[t]
\centering
  \footnotesize
  \renewcommand\arraystretch{0.8}
  \caption{Accuracy (mean±standard deviation) comparisons between the existing neural architectures and BiaslessNAS using the Top-5 models trained by each neural architecture, in terms of highest reward in Eq. \ref{equ:reward}}
\begin{tabular}{ccccccc}
\hline
Model                                                                & Light Acc.(\%)      & Dark Acc.(\%)       & Overall(\%)    & Acc Imp.        & \begin{tabular}[c]{@{}c@{}}Unfair.\\ Score\end{tabular}        & Fair. Imp.    \\ \hline
MobilenetV2                                                          & 81.90$\pm$0.78          & 59.26$\pm$1.2           & 81.69$\pm$0.77          & baseline        & \begin{tabular}[c]{@{}c@{}}0.2264\\ $\pm$0.0194\end{tabular}          & baseline         \\
Resnet18                                                             & 82.54$\pm$1.48          & 63.59$\pm$1.14          & 82.36$\pm$1.47          & 0.67\% $\uparrow$          & \begin{tabular}[c]{@{}c@{}}0.1894\\ $\pm$0.0233\end{tabular}          & 16.34\% $\uparrow$          \\
ResNet34                                                             & 82.95$\pm$0.69          & 67.18$\pm$1.14          & 82.81$\pm$0.67          & 1.12\% $\uparrow$          & \begin{tabular}[c]{@{}c@{}}0.1577\\ $\pm$0.0181\end{tabular}          & 30.34\% $\uparrow$          \\
MnasNet                                                              & 76.54$\pm$1.20          & 61.02$\pm$3.34          & 76.40$\pm$1.22          & 5.29\%$\downarrow$           & \begin{tabular}[c]{@{}c@{}}0.1551\\ $\pm$0.0253\end{tabular}          & 31.49\% $\uparrow$          \\
\textbf{\begin{tabular}[c]{@{}c@{}}Biasless\\ NAS-Fair\end{tabular}} & \textbf{79.58$\pm$0.18} & \textbf{71.79$\pm$2.57} & \textbf{79.51$\pm$0.20} & \textbf{2.18\%}$\downarrow$  & \textbf{\begin{tabular}[c]{@{}c@{}}0.0779\\ $\pm$0.0252\end{tabular}} & \textbf{65.59\%}$\uparrow$  \\
\textbf{\begin{tabular}[c]{@{}c@{}}Biasless\\ NAS-Acc\end{tabular}}  & \textbf{84.37$\pm$0.53} & \textbf{69.23$\pm$1.81} & \textbf{84.24$\pm$0.52} & \textbf{2.55\%}$\uparrow$  & \textbf{\begin{tabular}[c]{@{}c@{}}0.1514\\ $\pm$0.0226\end{tabular}} & \textbf{33.13\%}$\uparrow$  \\ \hline
\end{tabular}
\label{tab:nasres}%

\end{table*}

\noindent\textbf{\textit{Evaluation of BiaslessNAS.}}
Table \ref{tab:nasres} reports the evaluation results. 
These two architectures were obtained from BiaslessNAS with the lowest unfairness score and the highest accuracy, respectively.
Two hyperparameters are used in the framework: (1) Alpha is the scalable parameter for accuracy, and (2) Beta is for fairness. We explore two settings: BiaslessNAS-Fair has a larger Beta (0.8) and a smaller Alpha (0.2), while BiaslessNAS-Acc has a larger Alpha (0.8) and a smaller Beta (0.2).
For a fair comparison of different neural architectures ($N$), all competitors are trained using the proposed fairness-aware data processing ($D$) and trainer ($f^{\prime}$).
As shown in Table \ref{tab:nasres}, it is clear that BiaslessNAS-Fair can achieve competitive accuracy with the lowest unfairness score over others. More specifically, the unfairness score of BiaslessNAS-Fair is only 0.0779 on average, which achieves an improvement of 65.59\% compared with MobileNetV2 regarding fairness.
On the other hand, BiaslessNAS-Acc achieves the highest accuracy with the lowest unfairness score against other existing models. 

\begin{figure}[t]
  \centering
  \includegraphics[width=1\textwidth]{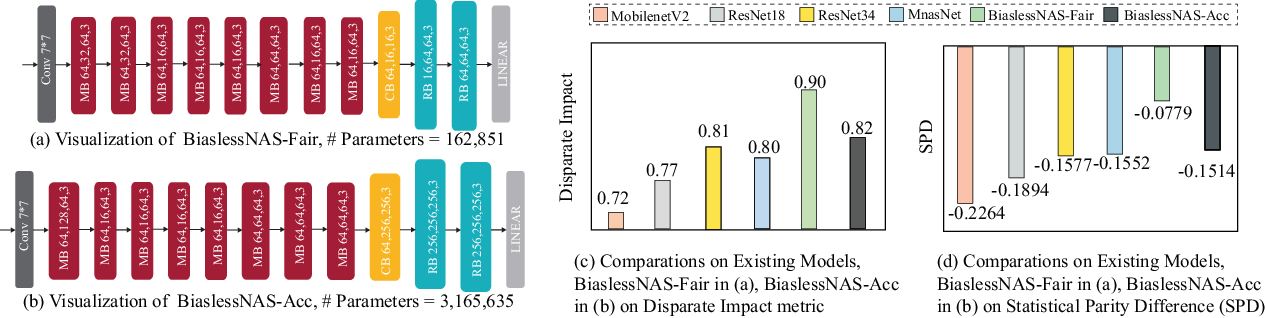}
  \caption{Visualization of BiaslessNAS-Fair and BiaslessNAS-Acc, together with their performance on different fairness metrics}
    
  \label{fig:DI}
  \label{fig:biasless}
  
\end{figure}

\noindent\textbf{\textit{Neural Architecture Visualization.}}
Fig. \ref{fig:biasless}(a)-(b) 
showcase the neural architectures derived from BiaslessNAS, highlighting the structural nuances between BiaslessNAS-Fair and BiaslessNAS-Acc. Despite sharing identical block types across layers, these architectures differ in the number of channels employed. Notably, both incorporate a MobileNet block at the outset for initial feature processing, followed by denser conventional and Residual blocks tailored to manage diverse group features. This visualization underscores the impact of neural architecture on fairness and suggests that strategically varying block types, particularly at the beginning and end of the architecture, can synergistically enhance fairness outcomes. This observation supports the premise that thoughtful architectural design is crucial in developing fair and effective architectures.


\noindent\textbf{\textit{BiaslessNAS is Fairer on Different Metrics.}}
In addition to the unfairness score defined in Equation \ref{equ:fair}, we further evaluate BiaslessNAS on other two commonly used fairness metrics: Disparate impact (DI) \cite{feldman2015certifying} and Statistical Parity Difference (SPD) \cite{li2021estimating}.
Fig. \ref{fig:DI}(c)-(d) present a comparison. In Fig. \ref{fig:DI}(c), BiaslessNAS-Fair stands out by achieving the highest DI value, indicating its superiority in fairness over other examined architectures. Fig. \ref{fig:DI}(d) reveals that models with SPD scores closer to zero are preferable, with BiaslessNAS-Fair and BiaslessNAS-Acc emerging as the top performers in this regard. These findings collectively demonstrate that BiaslessNAS effectively identifies solutions that surpass conventional neural architectures in fairness across different metrics.


\begin{figure}[t]
\centering
\includegraphics[width=1\textwidth]{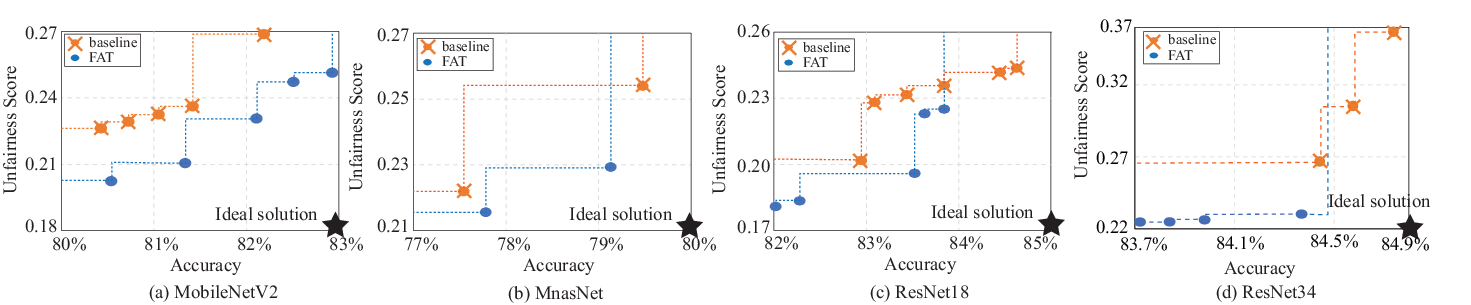}
\caption{Evaluation of fairness-aware trainer on the existing neural architectures} \label{fig:fairnessTrainer}

\end{figure}

\noindent\textbf{\textit{Evaluation of fairness-aware trainer:}} 
This ablation study is conducted by fixing the same $N\&D$ and comparing results for different $f^{\prime}$.
Fig. \ref{fig:fairnessTrainer} shows the evaluation results of the fairness-aware trainer on 4 existing neural architectures.
The baseline for each architecture has the setting of $\frac{o_l}{o_d}=\frac{|D_{g_l}|}{|D_{g_d}|}$, which means that the batch generator will use the same ratio between the number of dark-skin and light-skin images to load data.
On the other hand, the fairness-aware trainer (denoted as $FAT$) changes the ratio of $\frac{o_l}{o_d}$ to be 1.

In these figures, 
each dot is associated with one solution: the dots with a cross represent the baseline approach and the dots represent the FAT approach.
From the results in Fig. \ref{fig:fairnessTrainer}, we have several observations. (1) FAT can find neural architectures with lower unfairness scores.
(2) But, if the design is to maximize accuracy regardless of the fairness, then the baseline performs better than FAT (note that one exception is MobileNetV2, in which FAT dominates the baseline approach).
More specifically, when we compare the fairest architectures (i.e., the left-most dots for each approach in Fig. \ref{fig:fairnessTrainer}), FAT can achieve a 10.52\%, 50.20\%, 36.98\%, and 37.82\% reduction in unfairness scores on each architecture.
The above results clearly show that with the same neural architecture and data augmentation, the fairness-aware trainer can indeed improve fairness but it should be careful about the possible accuracy degradation.

\begin{table}[t]
  \centering
  \footnotesize
  \renewcommand\arraystretch{1}
  \caption{Quantitative Analysis of Three Fairness-related Factors on MobileNetV2}
  {
\begin{tabular}{ccccc}
\hline
Models & Acc. & Unfairness & DI & Ranking \\ \hline
\begin{tabular}[c]{@{}c@{}}MobilenetV2 (Vanilla)\end{tabular} & 81.05\% & 0.2325 & 0.71 & 5 \\ 
{\begin{tabular}[c]{@{}c@{}}MobilenetV2 with $f^\prime$\end{tabular}} & 81.34\% & 0.2105 & 0.74 & 4 \\ 
\begin{tabular}[c]{@{}c@{}}MobilenetV2 with ($D$ + $f^\prime$)\end{tabular} & 82.14\% & 0.1528 & 0.81 & 2 \\ 
\begin{tabular}[c]{@{}c@{}}FairNAS with $N$\end{tabular} \cite{sheng2022larger} & 84.06\% & 0.1755 & 0.79 & 3 \\ 
\textbf{\begin{tabular}[c]{@{}c@{}}BiaslessNAS-Acc with ($D$ + $f^\prime$ + $N$)\end{tabular}} & \textbf{84.24\%} & \textbf{0.1514} & \textbf{0.82} & \textbf{1} \\ \hline
\end{tabular}
    }
  \label{tab:table3}%

\end{table}%

\noindent\textbf{\textit{Evaluation of different optimization combinations.}}
This ablation study evaluates various optimization combinations to assess the benefits of co-optimize. The results, summarized in 
Table \ref{tab:table3}, contrast different strategies against a baseline MobileNetV2 architecture. Initially, we examine MobileNetV2 in its standard form, followed by versions enhanced with a fairness-aware trainer (denoted as 
$f^{\prime}$) and then with both a co-optimized trainer and data augmentation ($D+f^{\prime}$). The outcomes illustrate that co-optimization significantly enhances the fairness of MobileNetV2, as indicated by improvements in unfairness scores and disparate impact metrics.
In a further analysis, a fairness-aware Neural Architecture Search (NAS), termed "FairNAS," is introduced. FairNAS seeks to identify fair neural architectures without incorporating a fairness-aware trainer or data augmentation. Interestingly, FairNAS surpasses the fairness metrics of MobileNetV2 paired with $f^{\prime}$ alone but falls short of the combination of MobileNetV2 with $D+f^{\prime}$ in fairness metrics, albeit with a slight advantage in accuracy.
Introducing BiaslessNAS-Acc, which integrates data-algorithm-architecture ($D+f^{\prime}+N$) reveals that this approach outperforms FairNAS by achieving higher accuracy and further enhancing fairness. This comprehensive co-optimization of data, algorithm, and architecture emerges as the most effective strategy, showcasing the superior efficacy of simultaneous optimization across these dimensions for advancing both accuracy and fairness in machine learning models.


The above results give us the following three insights. (1) Neural architecture indeed affects fairness. It can even make a larger impact on fairness than the fairness-aware trainer. (2) The neural architecture search is good at identifying architectures with high accuracy. But without the help of a fairness-aware trainer and data augmentation, it may not optimize the fairness in the search loop. (3) Co-optimization is essential to make the best accuracy-fairness tradeoff.

\section{Conclusion}


In this paper, we delve into the factors influencing fairness in ML systems, unveiling that optimizing models, algorithms, and data collectively can better balance accuracy and fairness. We introduce a novel framework, BiaslessNAS, designed for this holistic optimization approach, specifically targeting the inherent biases in skin lesion datasets. To ensure accuracy and fairness, BiaslessNAS incorporates a fairness-aware training mechanism that creates balanced data batches and refines weighted loss to enhance the fairness of minority groups. Additionally, a reinforcement learning optimizer steers the co-optimization process, proving that this integrated approach markedly surpasses traditional methods that optimize data, algorithms, and architecture separately. Our evaluations confirm that co-optimization significantly enhances fairness without compromising accuracy.


\subsubsection{\ackname}  
We gratefully acknowledge the support of the National Institutes of Health (NIH) (Award No. 1R01EB033387-01).

\subsubsection{\discintname}
The authors have no competing interests to declare that are relevant to the content of this article.










%
%

\clearpage
\scriptsize
\bibliographystyle{splncs04}
\bibliography{paper.bib}

\begin{thebibliography}{10}
\providecommand{\url}[1]{\texttt{#1}}
\providecommand{\urlprefix}{URL }
\providecommand{\doi}[1]{https://doi.org/#1}

\bibitem{Altas}
Dermatology atlas. \url{http://www.atlasdermatologico.com.br/}, accessed Nov, 2021

\bibitem{Dermnet}
Dermnet dataset. \url{http://www.dermnet.com/}, accessed Nov, 2021

\bibitem{esfair}
Fair and intelligent embedded system challenge at esweek 2023. \url{https://esfair2023.github.io/ESFair/Submission.html}

\bibitem{biasapp}
Gender and skin-type bias in commercial ai systems. \url{https://news.mit.edu/2018/study-finds-gender-skin-type-bias-artificial-intelligence-systems-0212}

\bibitem{ISIC2019}
Skin lesion analysis. \url{https://challenge2019.isic-archive.com/}

\bibitem{abusitta2019generative}
Abusitta, A., A{\"\i}meur, E., Wahab, O.A.: Generative adversarial networks for mitigating biases in machine learning systems. arXiv preprint arXiv:1905.09972  (2019)

\bibitem{bahng2020learning}
Bahng, H., Chun, S., Yun, S., Choo, J., Oh, S.J.: Learning de-biased representations with biased representations. In: International Conference on Machine Learning. pp. 528--539. PMLR (2020)

\bibitem{chiu2023toward}
Chiu, C.H., Chung, H.W., Chen, Y.J., Shi, Y., Ho, T.Y.: Toward fairness through fair multi-exit framework for dermatological disease diagnosis. In: International Conference on Medical Image Computing and Computer-Assisted Intervention. pp. 97--107. Springer (2023)

\bibitem{de2020use}
De, A., Sarda, A., Gupta, S., Das, S.: Use of artificial intelligence in dermatology. Indian journal of dermatology  \textbf{65}(5), ~352 (2020)

\bibitem{feldman2015certifying}
Feldman, M., Friedler, S.A., Moeller, J., Scheidegger, C., Venkatasubramanian, S.: Certifying and removing disparate impact. In: proceedings of the 21th ACM SIGKDD international conference on knowledge discovery and data mining. pp. 259--268 (2015)

\bibitem{hao2021towards}
Hao, W., El-Khamy, M., Lee, J., Zhang, J., Liang, K.J., Chen, C., Duke, L.C.: Towards fair federated learning with zero-shot data augmentation. In: Proceedings of the IEEE/CVF Conference on Computer Vision and Pattern Recognition. pp. 3310--3319 (2021)

\bibitem{he2016deep}
He, K., Zhang, X., Ren, S., Sun, J.: Deep residual learning for image recognition. In: Proceedings of the IEEE conference on computer vision and pattern recognition. pp. 770--778 (2016)

\bibitem{howard2017mobilenets}
Howard, A.G., Zhu, M., Chen, B., Kalenichenko, D., Wang, W., Weyand, T., Andreetto, M., Adam, H.: Mobilenets: Efficient convolutional neural networks for mobile vision applications. arXiv preprint arXiv:1704.04861  (2017)

\bibitem{jiang2020hardware}
Jiang, W., Yang, L., Sha, E.H.M., Zhuge, Q., Gu, S., Dasgupta, S., Shi, Y., Hu, J.: Hardware/software co-exploration of neural architectures. IEEE Transactions on Computer-Aided Design of Integrated Circuits and Systems  \textbf{39}(12),  4805--4815 (2020)

\bibitem{jiang2019accuracy}
Jiang, W., Zhang, X., Sha, E.H.M., Yang, L., Zhuge, Q., Shi, Y., Hu, J.: Accuracy vs. efficiency: Achieving both through fpga-implementation aware neural architecture search. In: Proceedings of the 56th Annual Design Automation Conference 2019. pp.~1--6 (2019)

\bibitem{jin2024learning}
Jin, C., Che, T., Peng, H., Li, Y., Pavone, M.: Learning from teaching regularization: Generalizable correlations should be easy to imitate. arXiv preprint arXiv:2402.02769  (2024)

\bibitem{kamulegeya2019using}
Kamulegeya, L.H., Okello, M., Bwanika, J.M., Musinguzi, D., Lubega, W., Rusoke, D., Nassiwa, F., B{\"o}rve, A.: Using artificial intelligence on dermatology conditions in uganda: A case for diversity in training data sets for machine learning. BioRxiv p. 826057 (2019)

\bibitem{krizhevsky2012imagenet}
Krizhevsky, A., Sutskever, I., Hinton, G.E.: Imagenet classification with deep convolutional neural networks. Advances in neural information processing systems  \textbf{25} (2012)

\bibitem{li2021estimating}
Li, X., Cui, Z., Wu, Y., Gu, L., Harada, T.: Estimating and improving fairness with adversarial learning. arXiv preprint arXiv:2103.04243  (2021)

\bibitem{miranda2022debiasing}
Miranda, T.C., Gimenez, P.F., Lalande, J.F., Tong, V.V.T., Wilke, P.: Debiasing android malware datasets: How can i trust your results if your dataset is biased? IEEE Transactions on Information Forensics and Security  \textbf{17},  2182--2197 (2022)

\bibitem{nakajima2019generating}
Nakajima, S., Chen, T.Y.: Generating biased dataset for metamorphic testing of machine learning programs. In: IFIP International Conference on Testing Software and Systems. pp. 56--64. Springer (2019)

\bibitem{nam2020learning}
Nam, J., Cha, H., Ahn, S., Lee, J., Shin, J.: Learning from failure: De-biasing classifier from biased classifier. Advances in Neural Information Processing Systems  \textbf{33},  20673--20684 (2020)

\bibitem{9488320}
Ouyang, N., Huang, Q., Li, P., Cai, Y., Liu, B., Leung, H.f., Li, Q.: Suppressing biased samples for robust vqa. IEEE Transactions on Multimedia  \textbf{24},  3405--3415 (2022). \doi{10.1109/TMM.2021.3097502}

\bibitem{peng2023autorep}
Peng, H., Huang, S., Zhou, T., Luo, Y., Wang, C., Wang, Z., Zhao, J., Xie, X., Li, A., Geng, T., et~al.: Autorep: Automatic relu replacement for fast private network inference. In: Proceedings of the IEEE/CVF International Conference on Computer Vision. pp. 5178--5188 (2023)

\bibitem{peng2024lingcn}
Peng, H., Ran, R., Luo, Y., Zhao, J., Huang, S., Thorat, K., Geng, T., Wang, C., Xu, X., Wen, W., et~al.: Lingcn: Structural linearized graph convolutional network for homomorphically encrypted inference. Advances in Neural Information Processing Systems  \textbf{36} (2024)

\bibitem{peng2024maxk}
Peng, H., Xie, X., Shivdikar, K., Hasan, M.A., Zhao, J., Huang, S., Khan, O., Kaeli, D., Ding, C.: Maxk-gnn: Extremely fast gpu kernel design for accelerating graph neural networks training. In: Proceedings of the 29th ACM International Conference on Architectural Support for Programming Languages and Operating Systems, Volume 2. pp. 683--698 (2024)

\bibitem{sandler2018mobilenetv2}
Sandler, M., Howard, A., Zhu, M., Zhmoginov, A., Chen, L.C.: Mobilenetv2: Inverted residuals and linear bottlenecks. In: Proc. of CVPR. pp. 4510--4520 (2018)

\bibitem{shafahi2019adversarial}
Shafahi, A., Najibi, M., Ghiasi, M.A., Xu, Z., Dickerson, J., Studer, C., Davis, L.S., Taylor, G., Goldstein, T.: Adversarial training for free! Advances in Neural Information Processing Systems  \textbf{32} (2019)

\bibitem{sharma2020data}
Sharma, S., Zhang, Y., R{\'\i}os~Aliaga, J.M., Bouneffouf, D., Muthusamy, V., Varshney, K.R.: Data augmentation for discrimination prevention and bias disambiguation. In: Proceedings of the AAAI/ACM Conference on AI, Ethics, and Society. pp. 358--364 (2020)

\bibitem{sheng2022larger}
Sheng, Y., Yang, J., Wu, Y., Mao, K., Shi, Y., Hu, J., Jiang, W., Yang, L.: The larger the fairer? small neural networks can achieve fairness for edge devices. In: Proceedings of the 59th ACM/IEEE Design Automation Conference. pp. 163--168 (2022)

\bibitem{simonyan2014very}
Simonyan, K., Zisserman, A.: Very deep convolutional networks for large-scale image recognition. arXiv preprint arXiv:1409.1556  (2014)

\bibitem{spinde2021towards}
Spinde, T., Krieger, D., Plank, M., Gipp, B.: Towards a reliable ground-truth for biased language detection. In: 2021 ACM/IEEE Joint Conference on Digital Libraries (JCDL). pp. 324--325. IEEE (2021)

\bibitem{targ2016resnet}
Targ, S., Almeida, D., Lyman, K.: Resnet in resnet: Generalizing residual architectures. arXiv preprint arXiv:1603.08029  (2016)

\bibitem{wang2020ica}
Wang, T., Xu, X., Xiong, J., Jia, Q., Yuan, H., Huang, M., Zhuang, J., Shi, Y.: Ica-unet: Ica inspired statistical unet for real-time 3d cardiac cine mri segmentation. In: International conference on medical image computing and computer-assisted intervention. pp. 447--457. Springer (2020)

\bibitem{williams1992simple}
Williams, R.J.: Simple statistical gradient-following algorithms for connectionist reinforcement learning. Machine learning  \textbf{8}(3-4),  229--256 (1992)

\bibitem{zheng2021hierarchical}
Zheng, H., Han, J., Wang, H., Yang, L., Zhao, Z., Wang, C., Chen, D.Z.: Hierarchical self-supervised learning for medical image segmentation based on multi-domain data aggregation. In: International Conference on Medical Image Computing and Computer-Assisted Intervention. pp. 622--632. Springer (2021)

\end{thebibliography}





\end{document}